\documentclass{article}

    \usepackage[nonatbib,final]{neurips_2021}

\usepackage[utf8]{inputenc} 
\usepackage[T1]{fontenc}    
\usepackage{hyperref}       
\usepackage{url}            
\usepackage{booktabs}       
\usepackage{amsfonts}       
\usepackage{nicefrac}       
\usepackage{microtype}      
\usepackage{xcolor}         
\usepackage{caption} 
\usepackage{graphicx}
\usepackage{booktabs} 
\usepackage{amssymb}
\usepackage{amsmath}
\usepackage{multirow}
\usepackage{subcaption}
\usepackage{float}
\usepackage{mathtools}
\usepackage{bbold}

\usepackage[linesnumbered,ruled,vlined]{algorithm2e}
\DontPrintSemicolon

\SetKwComment{Comment}{\color{green!50!black}// }{}

\newcommand{\var}{\texttt}

\SetKwProg{Function}{function}{}{}

\usepackage{pythontex} 
\usepackage{makecell}
\usepackage{wrapfig}

\title{Error Correction Code Transformer}

\author{%
  Yoni Choukroun \\
  Tel Aviv University\\
  \texttt{choukroun.yoni@gmail.com} \\ \\
  \And Lior Wolf \\
  Tel Aviv University\\
      \texttt{liorwolf@gmail.com} \\ \\
}

\begin{document}

\maketitle
\begin{abstract}
Error correction code is a major part of the communication physical layer, ensuring the reliable transfer of data over noisy channels.
Recently, neural decoders were shown to outperform classical decoding techniques.
However, the existing neural approaches present strong overfitting due to the exponential training complexity, or a restrictive inductive bias due to reliance on Belief Propagation.
Recently, Transformers have become methods of choice in many applications thanks to their ability to represent complex interactions between elements.
In this work, we propose to extend for the first time the Transformer architecture to the soft decoding of linear codes at arbitrary block lengths.
We encode each channel's output dimension to high dimension for better representation of the bits information to be processed separately.
The element-wise processing allows the analysis of the channel output reliability, while the algebraic code and the interaction between the bits are inserted into the model via an adapted masked self-attention module.
The proposed approach demonstrates the extreme power and flexibility of Transformers and outperforms existing state-of-the-art neural decoders by large margins at a fraction of their time complexity.
\end{abstract}
\section{Introduction}
Reliable digital communication is of major importance in the modern information age and involves the design of codes to be robustly decoded under noisy transmission channels.
While optimal decoding is defined by the NP-hard maximum likelihood rule, the efficient decoding of algebraic block codes is an open problem.
Recently, powerful learning-based techniques have been introduced to this field.

At the present time, deep learning models implementing parameterized versions of the legacy Belief Propagation (BP)  decoders dominate \cite{nachmani2016learning, nachmani2017learning,lugosch2017neural,nachmani2019hyper,buchberger2020learned}.
This approach is appealing since the exponential size of the input space is mitigated by ensuring that certain symmetry conditions are met. 
In this case, it is sufficient to train the network on a noisy version of the zero codeword.
However, this approach remains extremely restrictive due to the strong inductive bias induced by the corresponding Tanner graph. 

Model-free decoders \cite{AutoencoderComm,gruber2017deep,kim2018communication} employ generic neural networks and may potentially benefit from the application of powerful architectures that have emerged in recent years in various fields.
However, such decoders suffer from the curse of dimensionality, since the training grows exponentially with the number of information bits \cite{wang1996artificial}. 
A major challenge {of models free decoders}, therefore, is difficulty of the network to learn the code, and to detect most reliable components of the code, especially for the basic multi-layers perceptron based architectures suggested in the literature. 

In this work, we propose a \emph{model free} decoder built upon the Transformer architecture \cite{vaswani2017attention}. As far as we can ascertain, this is the first time Transformers is adapted to error correction codes. 
Our Transformer employs a high dimensional \emph{scaled} element-wise embedding of the input that can be seen as a positional and reliability encoding.
Our decoder employs an \emph{adapted-mask} self-attention mechanism, in which the interaction between bits follows the code's parity check matrix, {\color{black}enabling the incorporation of domain knowledge into the model.}

Applied to a wide variety of codes, our method outperforms the state-of-the-art learning-based solutions by very large margins even with very shallow architectures, enabling potential deployment. 
This is the first time a decoder designed de-novo outperforms neural BP-based decoders.

\section{Related Works}
Over the past few years, the emergence of deep learning has demonstrated the advantages of Neural Networks in many communication applications such as channel equalization, modulation, detection, quantization, compression, and \emph{decoding} \cite{ibnkahla2000applications}. 
In \cite{AutoencoderComm} the entire coding-decoding channel transmission pipeline was abstracted as a fully parameterized autoencoder.
In many scenarios, neural networks based methods outperform existing decoders, as well as popular codes \cite{nachmani2017learning,gruber2017deep}.

Two main classes of Neural decoders can be established in the current literature \cite{raviv2020graph}. 
Model-free decoders employ general neural network architectures as in \cite{cammerer2017scaling,gruber2017deep,kim2018communication,bennatan2018deep}.
The results obtained in \cite{gruber2017deep} for polar codes ($n=16$) are similar to the maximum a posteriori (MAP) decoding results.
However, the exponential number of possible codewords make the decoding of larger codes infeasible.
In \cite{bennatan2018deep}, a preprocessing of the channel output allows the decoder to remain provably invariant to the codeword and enables the deployment of any kind of model without any overfitting cost.
The Model-free approaches generally make use of stacked fully connected networks or recurrent neural networks in order to simulate the iterative process existing in many legacy decoders.
However, these architectures have difficulties in learning the code and analyzing the reliability of the output, and generally require prohibitively huge parameterization or expensive graph permutation preprocessing \cite{bennatan2018deep}.

The second class is that of  model-based decoders 
\cite{nachmani2016learning,nachmani2017learning} implementing parameterized versions of classical BP decoders where the Tanner graph is unfolded into a neural network in which weights are assigned to each variable edge, showing improvement in comparison to the baseline BP method.
Recently, \cite{raviv2020graph} presented a data-driven framework for codeword permutation selection as a pre-decoding stage.
A self-attention mechanism is applied over several permutations embedding to estimate the best permutation (among a given set) to be decoded.
 A  combination of the BP model with hyper-graph network was shown to enhance performance~\cite{nachmani2019hyper}. Further improvements over the latter were obtained by introducing an autoregressive signal based on previous state conditioning and SNR estimation~\cite{nachmani2021autoregressive}.

Transformer neural networks were originally introduced for machine translation \cite{vaswani2017attention} and they now dominate the performance tables for most applications in the field of Natural Language Processing (NLP), e.g., {\color{black}\cite{vaswani2017attention,devlin2018bert}}.  
Transformer encoders primarily rely on the self-attention operation in conjunction with feed-forward layers, allowing manipulation of variable-size sequences and learning of long-range dependencies.
Many works extended the Transformer architecture from NLP to various applications such as computer vision \cite{dosovitskiy2020image}, speech analysis \cite{liu2021tera} and molecular properties prediction \cite{choukroun2021geometric}.

\section{Background}
We provide the necessary background on error correction coding and the Transformer architecture.

\subsection{Coding}
We assume a standard transmission that uses a linear code $C$. The code is defined by the binary generator matrix $G$ of size $k \times n$ and the binary parity check matrix $H$ of size $(n - k) \times n$ defined such that $GH^{T}=0$ {\color{black} over the two elements Galois field}.

The input message $m \in \{0, 1\}^{k}$ is encoded by $G$ to a codeword $x \in C \subset \{0, 1\}^{n}$ satisfying $Hx=0$ and transmitted via a Binary-Input Symmetric-Output channel (e.g. an AWGN channel).
Let $y$ denote the channel output represented as $y=x_{s}+z$ where $x_s$ denotes the Binary Phase Shift Keying (BPSK) modulation of $x$ (i.e., over $\{\pm 1\}$), and $z$ is a random noise which is independent of the transmitted $x$.

\begin{figure}[t]
\centering
\includegraphics[width=1\columnwidth]{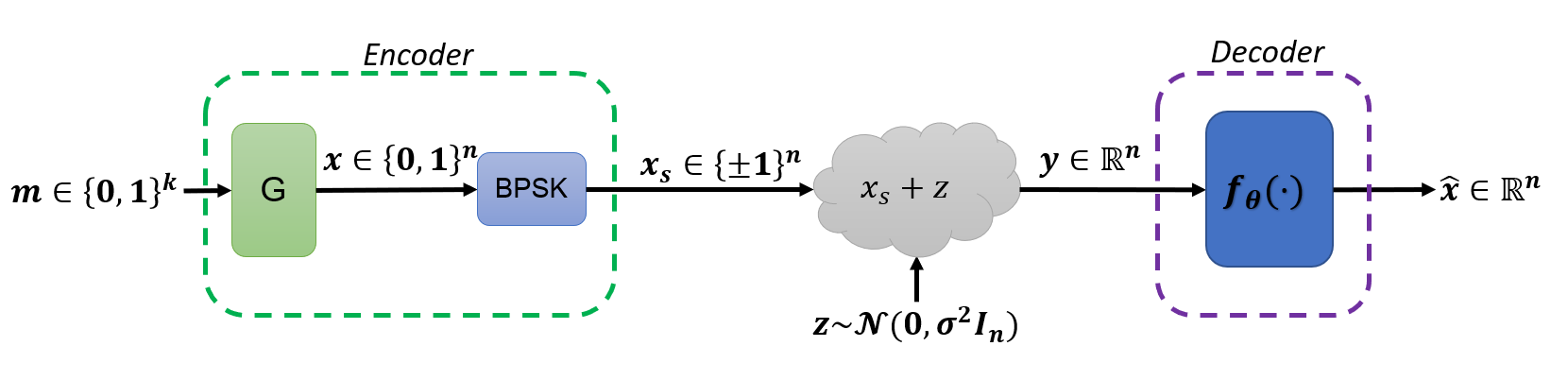}
\caption{Illustration of the communication system. Our work focus on the design and training of the parameterized decoder $f_{\theta}$.}
\label{fig:ecc_illustration}
\end{figure}

The main goal of the decoder {\color{black}$f:\mathbb{R}^{n}\rightarrow \mathbb{R}^{n}$} is to provide a soft approximation $\hat{x}=f(y)$ of the codeword. An illustration of the coding framework is presented in Figure \ref{fig:ecc_illustration}.

We wish to employ the Transformer architecture as part of a model-free approach for decoding. We, therefore, follow the preprocessing and post-processing of \cite{bennatan2018deep} to remain provably invariant to the transmitted codeword and to avoid overfitting. 
{This step involves a transformation of the channel output $y$ entailing a loss of information without implying any intrinsic performance penalty in decoding, i.e., the preprocessing coupled with an appropriately designed decoder can achieve MMSE decoding.} 

The preprocessing replaces $y$ with a vector of a dimensionality $2n-k$ defined as
\begin{equation}
\tilde{y} = h(y) = [|y|,s(y)]\,,
\end{equation}
where, $[\cdot ,\cdot]$ denotes vectors concatenation, $|y|$ denotes the absolute value (magnitude) of $y$ and $s(y)=Hy_{b}\in \{0, 1\}^{n-k}$ denotes the binary code \emph{syndrome} obtained via the multiplication of the binary mapping $y_b$ of $y$ by the parity check matrix.

The post-processing step plugs back the vector elements of $y$. Namely, the prediction takes the form
\begin{equation}
\hat{x}=y\cdot f(h(y))\,.
\end{equation}
In our setting the decoder is parameterized by $\theta$ and our model takes the explicit form:
\mbox{$\hat{x}=y\cdot f_{\theta}(|y|,Hy_{b})$}.

\subsection{Transformers}
The Transformer was introduced as a novel, attention-based building block for machine translation~\cite{vaswani2017attention}. The input sequence is first embedded into a high-dimensional space, coupled with positional embedding for each element. The embeddings are then propagated through multiple normalized self-attention and feed-forward blocks. 

The self-attention mechanism introduced by Transformers is based on a trainable associative memory with (key, value) vector pairs where a query vector $q \in \mathbb{R}^d$ is matched against a set of $k$ key vectors using scaled inner products as follows
\begin{equation}
\begin{aligned}
\label{transformer_att}
A(Q,K,V)=\text{Softmax}\bigg(\frac{QK^{T}}{\sqrt{d}}\bigg)V,
\end{aligned}
\end{equation}
where $Q \in \mathbb{R}^{N \times d}$, $K \in \mathbb{R}^{k \times d}$ and $V \in \mathbb{R}^{k \times d}$ represent the packed $N$ queries, $k$ keys and values tensors respectively.
Keys, queries and values are obtained using linear transformations of the sequence' elements.
A multi-head self-attention layer is defined by extending the self-attention mechanism using $h$ attention \emph{heads}, i.e. $h$ self-attention functions applied to the input, reprojected to values via a $dh \times D$ linear layer.
\section{Error Correction Code Transformer}
{We present the elements of the proposed Transformer decoder, the complete architecture, and the training procedure.}
\subsection{Positional Reliability Encoding}
We first consider each dimension of $\{\tilde{y}_{i}\}_{i=1}^{2n-k}$ separately and project each one to a high $d$ dimensional embedding $\{\phi_{i}\}_{i=1}^{2n-k}$ such that 
\begin{equation}
\begin{aligned}
\label{embedding}
  \phi_{i} =
    \begin{cases}
      |y_{i}|W_{i}, & \text{if} \ \ i\leq n\\
      \big(1-2(s(y))_{i-n+1}\big)W_{i}, & \text{otherwise}
    \end{cases}     
\end{aligned}
\end{equation}
where \mbox{$\{W_{j} \in \mathbb{R}^{d}\}_{j=1}^{2n-k}$} denotes the one-hot encoding defined according to the bit position. 
The embedding is modulated by the magnitude and the syndrome values such that less reliable elements (i.e. low magnitude) collapse to the origin.
This propriety becomes especially appealing when applied to the standard dot product of the self-attention module: for the first layer and two distinct information bits embedding $\phi_{i},\phi_{j}$ we have
\begin{equation}
\begin{aligned}
\label{dot_prod_reliability}
  \langle \phi_{i},\phi_{j} \rangle =
    \begin{cases}
      |y_{i}||y_{j}|\langle W_{i},W_{j} \rangle &  i,j\leq n\\
      |y_{i}|(1-2(s(y))_{j-n+1}\big)\langle W_{i},W_{j} \rangle, & i\leq n<j,
    \end{cases}   
\end{aligned}
\end{equation}
where an unreliable information bit vanishes and a non-zero syndrome entails a negative scaling, potentially less impacting the softmax aggregation as illustrated in Sec.~\ref{sec:discussion}.

In contrast to \cite{bennatan2018deep},  which requires permutation of the code in order to artificially provide to the fully connected network indications about the most reliable channel outputs, in our construction, the channel output reliability is directly obtained and maintained in the network via the scaled bit-wise embedding.
The proposed scaled encoding can be thought of as a \emph{positional} encoding according to the input \emph{reliability} since the bit positions are fixed. We note that representing the codeword by a set of one-hot vectors is very much different than any of the existing methods.

\subsection{Code Aware Self-Attention}
{In order to detect and correct errors, a decoder must \emph{analyze and compare} the received word bits via the parity check equations, such that a non-zero syndrome indicates that channel errors have occurred. In our Transformer based architecture, bits comparisons become natural via the self-attention \emph{interaction} mechanism.}

However, comparing every pair of elements as generally performed in Transformers architectures is sub-optimal, since not every bit is necessarily related to all the others. We propose to make use of the self-attention mechanism in order to incorporate fundamental domain knowledge about the relevant code. Specifically, it is used to indicate that the syndrome values should be dependent on the corresponding parity check bits solely.

\begin{algorithm}[b]
  \caption{Mask construction Pseudo Code}
  \Function{g(H)}{
    \var{l,n = H.shape}\\
    \var{k = n-l}\\
    \var{mask = eye(2n-k)}\\
    \For{i in range(0,n-k)}{
    \var{idx = where(H[i]==1)}
    \\
    \For{j in idx}{
    \var{mask[n+i,j] = mask[j,n+i] = 1}\\
    \For{k in idx}{
    \var{mask[j,k] = mask[k,j] = 1}
    }
    }
    }
    \Return{\var{$-\infty$($\lnot$ mask)}}
  }
  \label{algo:mask}
\end{algorithm}

Thus, for a given algebraic code defined by the matrix $H$ we propose to defined a function \mbox{$g(H):\{0,1\}^{(n-k)\times k}\rightarrow \{-\infty,0\}^{2n-k\times 2n-k}$ } defining the mask to be applied to the self-attention mechanism such that 
\begin{equation}
\begin{aligned}
\label{transformer_att_ecct}
A_{H}(Q,K,V)=\text{Softmax}\bigg(\frac{QK^{T}+ g(H)}{\sqrt{d}}\bigg)V.
\end{aligned}
\end{equation}

\begin{figure}[t]
\centering
\includegraphics[width=0.99\textwidth,trim={0 0 0 0}]{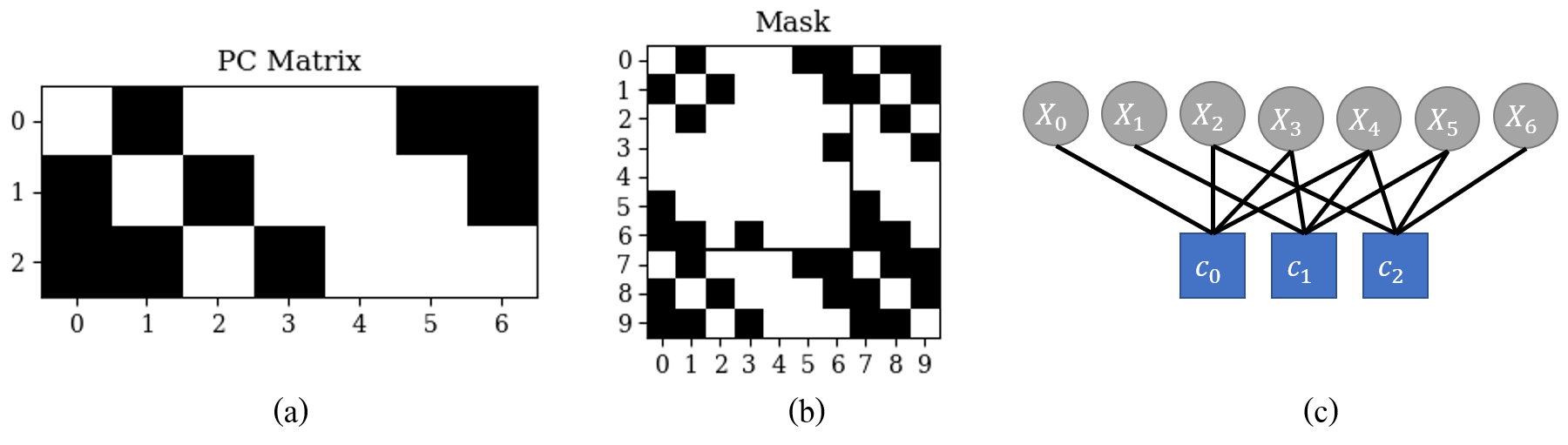}
\caption{We present the corresponding parity-check matrix (a), the induced Tanner graph (c) and the proposed masking (b) on the Hamming(7,4) code.}
\label{fig:mask_sample}
\end{figure}

We propose to build the \emph{symmetric} mask such that it contains information about \emph{every} pairwise bit relations as follows.
The mask is first initialized as the identity matrix.
For each row $i$ of the binary parity check matrix $H$, we unmask at the locations of \emph{every pair} of ones in the row, since those bits are connected and may impact each other as well as the syndrome in the decoding procedure.
We also unmask the location of pair of ones with the corresponding syndrome bit at $n+i$, since they define the parity check equations. 

We summarize the construction of the mask in Algorithm \ref{algo:mask}. This construction enables more freedom in decoding than the relations enabled by the Tanner graph, since related bits may have an impact on each other \emph{beyond} the parity check equations {as depicted in Figure \ref{fig:mask_sample}. 
While regular Transformer can be assimilated to a neural network applied on a complete graph, the proposed mask can be seen as the adjacency matrix of the Tanner graph extended to a two rings connectivity.
In contrast to BP, which collapses the information via interleaved variable and check layers, the masked self-attention allows the simultaneous cross-analysis of related elements.
} 

Most importantly, since the mask is fixed and computed once, the self-attention quadratic complexity bottleneck $\mathcal{O}(n^{2}d)$ is now reduced to the density of the code $\mathcal{O}(\sum_{ij}(H)_{ij}d)$.
This propriety is especially appealing for low density codes (e.g. \cite{gallager1962low}), while in all our experiments the original complexity of the self-attention is reduced by $84\%$ in average as presented in Sec.~\ref{sec:discussion}. 

\subsection{Architecture and Training}

\begin{figure}[t]
\centering
\includegraphics[width=1\textwidth]{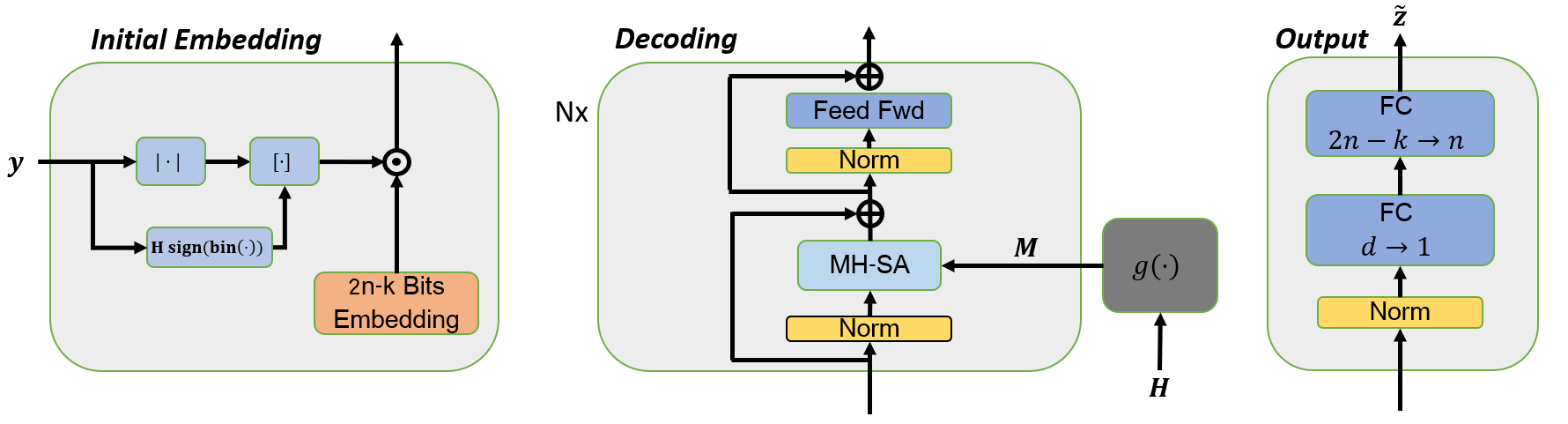}
\caption{Illustration of the proposed Transformer architecture.
The main differences from other Transformers are the initial scaled bit-embedding, the construction of the code aware masked self-attention, and the output module.}
\label{fig:ecct_arch}
\end{figure}

The initial encoding is defined as one hot encoding of the $2n-k$ code positions and set the model embedding dimension $d$.
The decoder is defined as a concatenation of $N$ decoding layers composed of self-attention and feed-forward layers interleaved by normalization layers.
The output module is defined by two fully connected layers.
The first layer reduces the element-wise embedding to a one-dimensional $2n-k$ vector and the second to a $n$ dimensional vector representing the soft decoded noise.
An illustration of the model is given in Figure \ref{fig:ecct_arch}.

The dimension of the feed-forward network is four times the size of the embedding dimension \cite{vaswani2017attention} and is composed of GEGLU layers \cite{shazeer2020glu} and layer normalization is set in pre-layer norm setting as in \cite{opennmt,xiong2020layer}.
We use an eight head self-attention module in all the experiments.

 The training objective is the  cross-entropy function where the goal is to learn to predict the \emph{multiplicative} noise $\tilde{z}$ \cite{bennatan2018deep}.
 Denoting the \emph{soft} multiplicative noise $\tilde{z}_{s}$ such that $y=x_{s}\tilde{z}_{s}$, we obtain $\tilde{z}_{s}=\tilde{z}_{s}x_{s}^{2}=yx_{s}$.
 Thus, the binary multiplicative noise to be predicted is defined by
 $\tilde{z}=\text{bin}(y\cdot x_{s})$, such that the loss computed for a single received word ${y}$ is
\begin{equation}
\begin{aligned}
\label{eq:loss}
\mathcal{L} = -\sum_{i=1}^{n} \tilde{z}_{i}\log(f_{\theta}(y))  +( 1-\tilde{z}_{i})\log(1-f_{\theta}(y)).
\end{aligned}
\end{equation}
The estimated \emph{binary} codeword is straightforwardly obtained as $\hat{x} = \text{bin}(\text{sign}(f_{\theta}(y)\cdot y)).$

{The Adam optimizer \cite{kingma2014adam} is used with 128 samples per minibatch for 1000 epochs with 1000 minibatches per epoch.
{\color{black}
For $N=10$ architectures we trained the models for 1500 epochs.
We note that more epochs can be beneficial for the performance. However, the current setting is more than enough to already set new SOTA performance.
By the model construction and its input preprocessing}, the zero codeword is enough for training.
The additive Gaussian noise is sampled randomly per batch in the $\{3,\dots,7\}$ normalized SNR (i.e. $E_b/N_0$) range.
We initialized the learning rate to $10^{-4}$ coupled with a cosine decay scheduler down to $5\cdot 10^{-7}$ at the end of the training. No warmup has been employed \cite{xiong2020layer}.}
\section[title]{Experiments\footnote{Code available at \url{https://github.com/yoniLc/ECCT}}}

In order to evaluate our method, we train the proposed architecture with three classes of linear block codes: Low-Density Parity Check (LDPC) codes~\cite{gallager1962low}, Polar codes~\cite{arikan2008channel} and Bose–Chaudhuri–Hocquenghem (BCH) codes~\cite{bose1960class}. 
All the parity check matrices are taken from~\cite{channelcodes}. 
The proposed architecture is solely defined by the number of encoders layers $N$ and the dimension of the embedding $d$.
We compare our method to the BP algorithm, to the augmented hypernetwork BP algorithm of \cite{nachmani2019hyper} (hyp BP), to the RNN architecture of \cite{bennatan2018deep} and the very recent SOTA performance of \cite{nachmani2021autoregressive} (AR BP).
All the results are obtained from the corresponding papers.
The results are reported as bit error rates (BER) for different normalized SNR values (dB). 
We follow the testing benchmark of \cite{nachmani2019hyper,nachmani2021autoregressive}. 
During test, our decoder decodes at least $10^{5}$ random codewords to obtain at least $500$ frames with errors at each SNR value.

The results are reported in Tab. ~\ref{tab:ber_snr} where we present the negative natural logarithm of the BER.
For each code we present the results of the BP based competing methods for 5 and 50 iterations (first and second row), corresponding to 10 and 100 layers neural network respectively.
We present our framework performance for six different architectures with $N=\{2,6\}$ $d=\{32,64,128\}$ respectively (first to third row).
Since LDPC codes are optimized for BP decoding \cite{richardson2001design}, we also present for these codes the performance of a $N=10,d=128$ architecture. 
We can see our approach outperforms the current SOTA results by very large margins on several codes at a fraction of the number of iterations and is able to outperform legacy methods (e.g. BP)  \emph{at convergence} even with extremely shallow architecture (e.g. $N=2,d=32$).
We remark the proposed method seems to perform better for high rate (i.e. $\frac{k}{n}$) codes.

\begin{table}[t]
    \centering
    \caption{A comparison of the negative natural logarithm of Bit Error Rate (BER) for three normalized SNR values of our method with literature baselines. 
	Higher is better. \\
	Concurrent results are obtained after $L=5$ BP iterations in first row (i.e. 10 layers neural network) and \emph{at convergence} results in second row obtained after $L=50$ BP iterations (i.e. 100 layers neural network).\\
	Best results in \textbf{bold}, second best is \underline{underlined}, {\color{black} and the minimal Transformer architecture to outperform every other competing method is in \textit{italic}.}\\
	Our performance are presented for seven different architectures: for $N=\{2,6\}$, we present results for $d=\{32,64,128\}$ (first to third row), and for $N=10$ we run only the $d=128$ configuration. }
    \label{tab:ber_snr}
    \resizebox{1\textwidth}{!}{%
    \begin{tabular}{lc@{~}c@{~}cc@{~}c@{~}cc@{~}c@{~}cc@{~}c@{~}cc@{~}c@{~}cc@{~}c@{~}c}
    \toprule
        Method & \multicolumn{3}{c}{BP\cite{pearl1988probabilistic}} &  \multicolumn{3}{c}{Hyp BP\cite{nachmani2019hyper}} & \multicolumn{3}{c}{AR BP\cite{nachmani2021autoregressive}} &
        \multicolumn{3}{c}{Ours N=2} & \multicolumn{3}{c}{Ours N=6} & \multicolumn{3}{c}{Ours N=10}\\
        \cmidrule(lr){2-4}
        \cmidrule(lr){5-7}
        \cmidrule(lr){8-10}
        \cmidrule(lr){11-13}
        \cmidrule(lr){14-16}
        \cmidrule(lr){17-19}
         & 4 & 5 & 6 & 4 & 5 & 6 & 4 & 5 & 6 & 4 & 5 & 6 & 4 & 5 & 6 & 4 & 5 & 6\\ 
         \midrule   
        Polar(64,32) 		& \makecell{3.52\\ 4.26} & \makecell{4.04\\ 5.38}	& \makecell{4.48\\ 6.50}	& \makecell{4.25\\ 4.59}  	& \makecell{5.49\\ 6.10}	& \makecell{7.02\\ 7.69}	& \makecell{4.77\\ 5.57 }   & \makecell{6.30\\7.43 }	& \makecell{8.19\\ 9.82 }		& \makecell{4.27\\ 4.57\\ 4.87} & \makecell{5.44\\ 5.86\\ 6.2} & \makecell{6.95\\ 7.50\\ 7.93} & \makecell{\textit{5.71}\\ \underline{6.48}\\ \bf6.99} & \makecell{\textit{7.63}\\ \underline{8.60}\\ \bf9.44} & \makecell{\textit{9.94}\\ \underline{11.43}\\ \bf12.32} \\
		 \midrule 																												
        Polar(64,48) 		& \makecell{4.15\\ 4.74} 	& \makecell{4.68\\ 5.94}	& \makecell{5.31\\ 7.42} & \makecell{4.91\\ 4.92}  	& \makecell{6.48\\ 6.44}	& \makecell{8.41\\ 8.39}	& \makecell{5.25\\ 5.41} 		 	& \makecell{6.96\\ 7.19}	& \makecell{9.00\\ 9.30} & \makecell{4.92\\ 5.14\\ 5.36 } & \makecell{6.46\\ 6.78\\ 7.12}  & \makecell{8.41\\ 8.9\\ \it9.39}  & \makecell{\it5.82\\ \underline{6.15}\\ \bf6.36}	&\makecell{\it7.81\\ \underline{8.20}\\ \bf8.46} & \makecell{10.24\\ \underline{10.86}\\ \bf11.09} \\			
         \midrule 
        Polar(128,64) 		& \makecell{3.38\\ 4.10} & \makecell{3.80\\ 5.11}	& \makecell{4.15\\ 6.15}	& \makecell{3.89\\ 4.52} 	& \makecell{5.18\\ 6.12}	& \makecell{6.94\\ 8.25}	& \makecell{4.02\\ 4.84}			& \makecell{5.48\\ 6.78}	& \makecell{7.55\\ 9.30}  & \makecell{3.51\\ 3.83\\ 4.04} & \makecell{4.52\\ 5.16\\ 5.52}  & \makecell{5.93\\ 7.04\\ 7.62}  & \makecell{4.47\\ \underline{\it5.12}\\ \bf5.92} & \makecell{6.34\\ \underline{\it7.36}\\ \bf8.64} & \makecell{8.89\\ \underline{\it10.48}\\ \bf12.18} \\
		 \midrule  						
        Polar(128,86) 		& \makecell{3.80\\ 4.49} & \makecell{4.19\\ 5.65}	& \makecell{4.62\\ 6.97}	& \makecell{4.57\\ 4.95}  & \makecell{6.18\\ 6.84}		& \makecell{8.27\\ 9.28}	& \makecell{4.81\\ 5.39} 	& \makecell{6.57\\ 7.37}		& \makecell{9.04\\ 10.13}	& \makecell{4.30\\ 4.49\\ 4.75}  & \makecell{5.58\\ 5.90\\ 6.25}  & \makecell{7.34\\ 7.75\\ 8.29}  & \makecell{5.36\\ \underline{\it5.75}\\ \bf6.31} & \makecell{\it7.45\\ \underline{8.16}\\ \bf9.01} & \makecell{\it10.22\\ \underline{11.29}\\ \bf12.45} \\
		 \midrule 							 				
        Polar(128,96) 		& \makecell{3.99\\ 4.61} & \makecell{4.41\\ 5.79}	& \makecell{4.78\\ 7.08}	& \makecell{4.73\\ 4.94}  	& \makecell{6.39\\ 6.76}	& \makecell{8.57\\ 9.09}	& \makecell{4.92\\ 5.27}	& \makecell{6.73\\ 7.44}		& \makecell{9.30\\ 10.2} 	& \makecell{4.56\\ 4.69\\ 4.88}  & \makecell{5.98\\ 6.20\\ 6.58}  & \makecell{7.93\\ 8.30\\ 8.93}  & \makecell{\it5.39\\ \underline{5.88}\\ \bf6.31} & \makecell{\it7.62\\ \underline{8.33}\\ \bf9.12} & \makecell{\it10.45\\ \underline{11.49}\\ \bf12.47} \\
		 \midrule 											
        LDPC(49,24) 		& \makecell{5.30\\ 6.23} & \makecell{7.28\\ 8.19}	& \makecell{9.88\\ 11.72}	& \makecell{5.76\\ 6.23} 	& \makecell{7.90\\ 8.54}	& \makecell{11.17\\ 11.95}		& \makecell{6.05\\ \bf6.58} 	& \makecell{8.13\\ \bf9.39}	& \makecell{11.68\\ 12.39}		& \makecell{4.33\\ 4.40\\ 4.53} & \makecell{5.79\\ 5.89\\ 6.10} & \makecell{7.69\\ 7.99\\ 8.33} & \makecell{ 5.45\\ 5.60\\ 5.79} & \makecell{7.65\\ 7.93\\ 8.13} & \makecell{10.58\\ 11.13\\ 11.40} & \underline{6.35} & \underline{9.01} & \textbf{\textit{12.43}} \\
		 \midrule 			
        LDPC(121,60) 		& \makecell{4.82\\ -} 	& \makecell{7.21\\ -}	& \makecell{10.87\\ -}	& \makecell{5.22\\ -} 		& \makecell{8.29\\ -}	& \makecell{13.00\\ -}	& \makecell{\underline{5.22}\\ -}	& \makecell{\underline{8.31}\\ -}		& \makecell{\underline{13.07}\\ -}			& \makecell{3.80\\ 3.81\\ 3.80} & \makecell{5.32\\ 5.36\\ 5.45} & \makecell{7.79\\ 7.87\\ 8.06} & \makecell{4.77\\ 4.88\\ 5.01} & \makecell{7.52\\ 7.77\\ 7.99} & \makecell{11.99\\ 12.39\\ 12.78} & \textbf{\textit{5.51}} & \textbf{\textit{8.89}} & \textbf{\textit{14.51}}\\
		 \midrule 													
        LDPC(121,70) 		& \makecell{5.88\\ -} 		& \makecell{8.76\\ -} & \makecell{13.04\\ -}	& \makecell{6.39\\ -}		& \makecell{9.81\\ -}	& \makecell{14.04\\ -}		& \makecell{\underline{6.45}\\ -} 	& \makecell{\underline{10.01}\\ -}		& \makecell{14.77\\ -} 		& \makecell{4.49\\ 4.52\\ 4.53} & \makecell{6.38\\ 6.44\\ 6.49} & \makecell{9.24\\ 9.36\\ 9.39} & \makecell{5.80\\ 6.04\\ 6.19} & \makecell{9.11\\ 9.54\\ 9.89} & \makecell{13.77\\ 14.65\\ \underline{\textit{15.58}}} & \textbf{\textit{6.86}} & \textbf{\textit{11.02}} & \textbf{{16.85}} \\
		 \midrule 											
        LDPC(121,80) 		& \makecell{6.66\\ -} 	& \makecell{9.82\\ -}	& \makecell{13.98\\ -}		& \makecell{6.95\\ -} & \makecell{10.68\\ -}		& \makecell{15.80\\ -}		& \makecell{\underline{7.22}\\-} 	& \makecell{\underline{11.03}\\-}		& \makecell{15.90\\-} 		& \makecell{5.11\\ 5.13\\ 5.24} & \makecell{7.18\\ 7.26\\ 7.46} & \makecell{10.18\\ 10.28\\ 10.66} & \makecell{6.63\\ 6.84\\ 7.07} & \makecell{10.28\\ 10.56\\ 10.96} & \makecell{15.10\\ 15.68\\ \underline{\textit{16.25}}}  & \textbf{\textit{7.76}} & \textbf{\textit{12.30}} & \textbf{{17.82}} \\
        \midrule 
        MacKay(96,48) 		& \makecell{6.84\\ -} 	& \makecell{9.40\\ -}	& \makecell{12.57\\ -}			& \makecell{7.19\\ -} 	& \makecell{10.02\\ -}		& \makecell{13.16\\ -}	& \makecell{\underline{7.43}\\ -}	& \makecell{\underline{10.65}\\ -}	& \makecell{\underline{14.65}\\ -}				& \makecell{4.92\\ 4.98\\ 5.12}  & \makecell{6.62\\ 6.72\\ 6.98}  & \makecell{8.88\\ 9.04\\ 9.44}  & \makecell{6.78\\ 7.02\\ 7.23} & \makecell{9.61\\ 10.12\\ 10.42} & \makecell{13.31\\ 14.18\\ 14.12} & \textbf{\textit{8.39}} & \textbf{\textit{12.24}} & \textbf{\textit{16.41}} \\
         \midrule 
        CCSDS(128,64) 		& \makecell{6.55\\ -} 	& \makecell{9.65\\ -}	& \makecell{13.78\\ -}		& \makecell{6.99\\ -} & \makecell{10.57\\ -}			& \makecell{15.27\\ -}		& \makecell{\underline{7.25}\\ -}  & \makecell{\underline{10.99}\\ -}		& \makecell{\underline{16.36}\\ -}			& \makecell{4.27\\ 4.39\\ 4.47} & \makecell{5.97\\ 6.03\\ 6.22} & \makecell{8.18\\ 8.43\\ 8.74} & \makecell{6.29\\ 6.49\\ 6.77} & \makecell{9.59\\ 10.18\\ 10.55} & \makecell{13.95\\ 14.76\\ 15.9} & \textbf{\textit{8.02}} & \textbf{\textit{12.60}} & \textbf{\textit{17.75}} \\
         \midrule 
        BCH(31,16) 			& \makecell{4.63\\ -} 	& \makecell{5.88\\ -}	& \makecell{7.60\\ -}	& \makecell{5.05\\ -}     & \makecell{6.64\\ -}     	& \makecell{8.80\\ -}	& \makecell{5.48\\ -}		& \makecell{7.37\\ -}		& \makecell{9.61\\ -}	& \makecell{4.51\\ 4.78\\ 5.18}  & \makecell{5.74\\ 6.15\\ 6.82}  & \makecell{7.35\\ 7.98\\ 8.91}   & \makecell{\it5.74 \\ \underline{5.85} \\ \bf6.39} & \makecell{\it7.42\\ \underline{7.52}\\ \bf8.29} & \makecell{\it9.59 \\ \underline{10.08}\\ \bf10.66} \\
         \midrule 
        BCH(63,36) & \makecell{3.72\\ 4.03} & \makecell{4.65\\ 5.42}	& \makecell{5.66\\ 7.26}	& \makecell{3.96\\ 4.29}    & \makecell{5.35\\ 5.91}    & \makecell{7.20\\ 8.01}	& \makecell{4.33 \\ 4.57} 	& \makecell{5.94\\ \underline{6.39}}	 	& \makecell{8.21\\ \underline{8.92}}	& \makecell{3.79\\ 4.05\\ 4.21} & \makecell{4.87\\ 5.28\\ 5.50} & \makecell{6.35\\ 7.01\\ 7.25} & \makecell{4.42\\ \underline{\textit{4.62}}\\ \bf4.86}	& \makecell{5.91\\ 6.24\\ \textbf{\textit{6.65}}}	& \makecell{8.01\\ 8.44\\ \textbf{\textit{9.10}}}	\\			
         \midrule 
        BCH(63,45) 			& \makecell{4.08\\ 4.36} & \makecell{4.96\\ 5.55}	& \makecell{6.07\\ 7.26}	& \makecell{4.48\\ 4.64} 		& \makecell{6.07\\ 6.27}	& \makecell{8.45\\ 8.51} & \makecell{4.80 \\ 4.97} 	& \makecell{6.43\\ 6.90}		& \makecell{8.69\\ 9.41}	& \makecell{4.47\\ 4.66\\ 4.79}   & \makecell{5.88\\ 6.16\\ 6.39} & \makecell{7.81\\ 8.17\\ 8.49} 	& \makecell{\it5.16\\ \underline{5.41}\\ \bf5.60} & \makecell{\it7.02\\ \underline{7.49}\\ \bf7.79}	& \makecell{\it9.75\\ \underline{10.25}\\ \bf10.93}	\\		
         \midrule 
        BCH(63,51)	 		& \makecell{4.34\\ 4.5} & \makecell{5.29\\ 5.82}	& \makecell{6.35\\ 7.42}	& \makecell{4.64\\ 4.80} 	& \makecell{6.08\\ 6.44}	& \makecell{8.16\\ 8.58}	& \makecell{4.95 \\ 5.17} 		& \makecell{6.69\\ 7.16}	& \makecell{9.18\\ 9.53}	& \makecell{4.60\\ 4.78\\ 5.01}  & \makecell{6.05\\ 6.34\\ 6.72}  & \makecell{8.05\\ 8.49\\ 9.03}  & \makecell{\it5.20\\  \underline{{5.46}}\\ \bf5.66}	& \makecell{7.08\\ \underline{\textit{7.57}}\\ \bf7.89}	& \makecell{\it9.65\\ \underline{{10.51}}\\ \bf11.01}	\\
		\bottomrule
	\end{tabular}
	}
\end{table}

\begin{figure}[t]
\vspace{1em}
\centering
\noindent  \begin{tabular}{@{}ccc@{}}
  \includegraphics[trim={0 0 0 0},clip, width=0.33\linewidth]{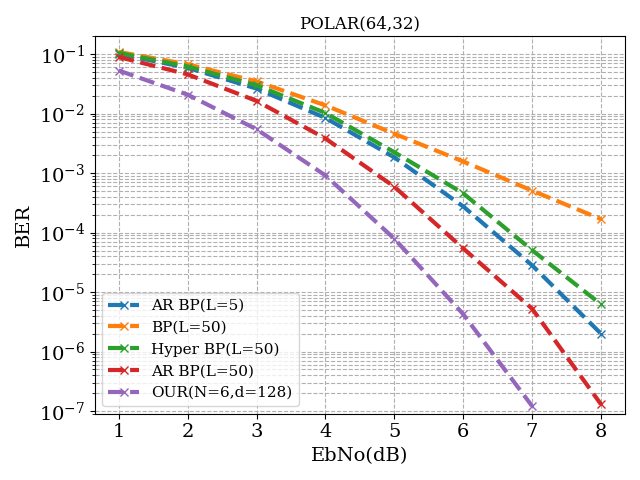}&
    \includegraphics[trim={0 0 0 0},clip, width=0.33\linewidth]{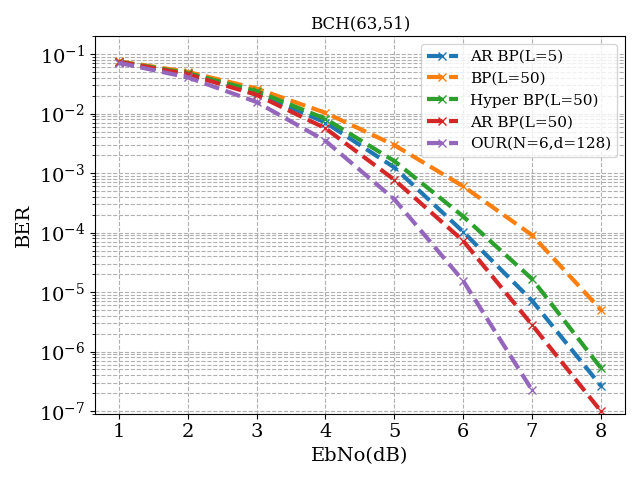}&
      \includegraphics[trim={0 0 0 0},clip, width=0.33\linewidth]{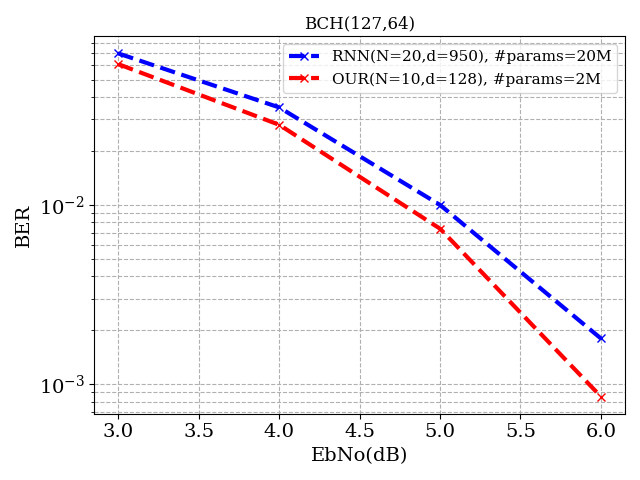}\\
      (a) & (b) & (c)\\
      \end{tabular}
  \caption{BER for various values of SNR for (a) Polar(64,32)  and (b) BCH(63,51) codes. Comparison with the best model of Bennatan et al. \protect\cite{bennatan2018deep} on BCH(127,64) code (c).}
\label{fig:BER_Results}
\end{figure}

We provide plots of more BER values for some of the codes in Figure \ref{fig:BER_Results}(a,b) for Polar(64,32)
and BCH(63,51) codes, respectively.
Our method is able to outperform the converged AR-BP model by up to two orders of magnitudes for high SNR values.

In Fig.~\ref{fig:BER_Results}(c), We compare our method with the best model of \cite{bennatan2018deep} on the only code provided BCH(127,64).
Their network is built as a five layers stacked GRU model with five iterations with an embedding of $5(2n-k)=950$ trained for 5000 epochs with a batch size of 1000. As can be seen, our method outperforms this model by up to $211\%$ ratio with only half the number of layers, one-tenth the number of parameters, and a fraction of the training requirements.


\section{Analysis}
\label{sec:discussion}
We study the impact of the proposed embedding and the masking procedure and we analyze and compare the complexity of our method to other methods.

\subsection{Impact of the Reliability Embedding} 
\begin{figure}[h]
\vspace{1em}
\centering
\includegraphics[width=1\textwidth]{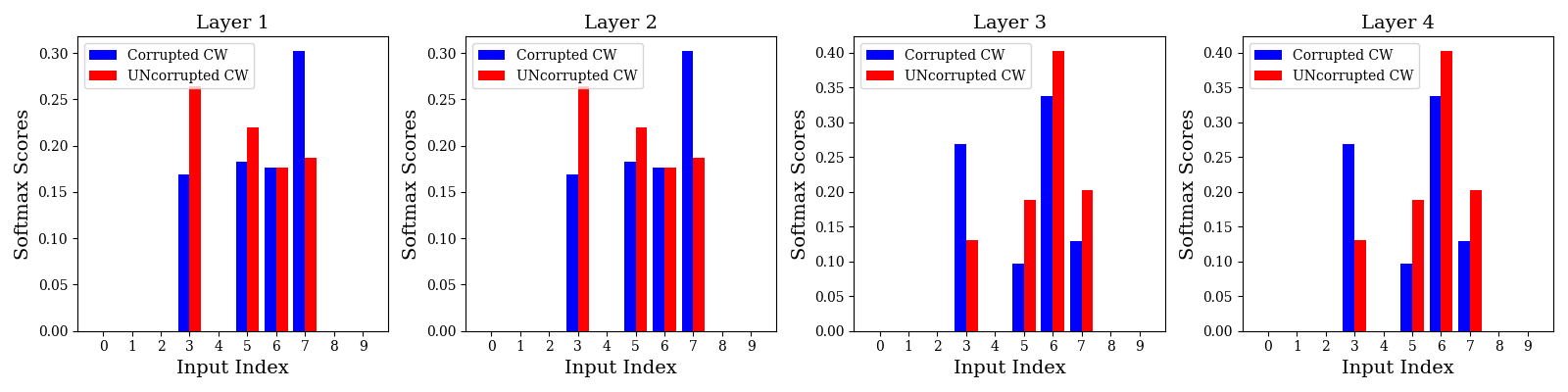}
\caption{Analysis of the scaled embedding values of the first column of the masked self-attention map with corrupted and uncorrupted codeword (CW) for the standardized Hamming code using a $(N,d)=(4,32)$ architecture. }
\label{fig:discuss_scaling}
\end{figure}

We depict in Figure \ref{fig:discuss_scaling} the impact of the \emph{scaled} embedding on the self-attention mechanism.
We chose the popular Hamming$(7,4)$ code where we corrupt the zero codeword with additive noise at the first bit (zero bit index), involving a non-zero syndrome at the first parity check equation.
We present the \emph{masked} softmax values of the self-attention map at index zero, i.e. the influence of the first (unreliable) bit on the self-attention aggregation (first column of the map).
We can see the first bit embedding is impactless when it is corrupted and then detectable, while its impact on the syndrome embedding is considerably increased.
When the network already corrected the bit (last layers(s)) the values return to normal.

\subsection{Ablation Study: Masking}
\begin{figure}
\centering
\begin{minipage}{.48\textwidth}
\centering
\includegraphics[width=0.9\textwidth]{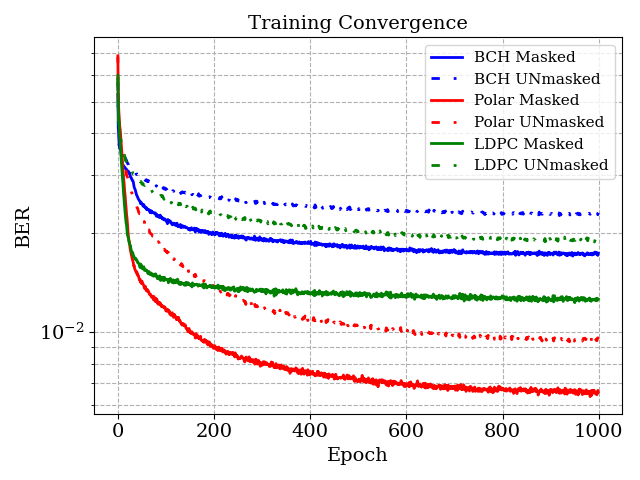}
\caption{Ablation study on  the proposed masking procedure on the training convergence for the BCH$(63,36)$, POLAR$(64,32)$ and LDPC$(49,24)$ using \mbox{$(N,d)=(6,64)$} architectures.
}
\label{fig:discuss_masking}
\end{minipage}%
\ \ \ \ 
\begin{minipage}{.48\textwidth}
\centering
\includegraphics[width=1\textwidth]{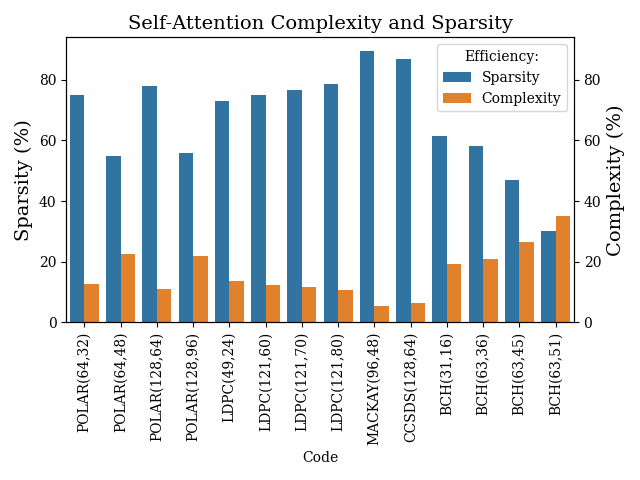}
\caption{Sparsity and complexity ratio of the proposed self-attention map for several codes. }
\label{fig:discuss_spars_comp}
\end{minipage}
\end{figure}

We present in Figure \ref{fig:discuss_masking} the impact of the proposed mask on the training convergence (overfitting cannot be noticed in our setting). 
We show an \emph{unmasked} architecture where we leave the network at learning the code by itself (i.e. $g(H)=0$), compared to our proposed masking approach.
{\color{black}
The benefit obtained with the mask is clear and can be understood by the  difficulties the Transformer may have in learning the code by itself.}
Therefore, the connectivity provided by the proposed masking framework reduces the loss by $76\%$, $69\%$ and $66\%$ for the BCH$(63,36)$, POLAR$(64,32)$ and LDPC$(49,24)$ codes respectively.


\subsection{Complexity Analysis}
We first present in Figure \ref{fig:discuss_spars_comp} the sparsity ratio of the \emph{masked} self-attention map, as well as the induced complexity ratio of the \emph{symmetric} map for several codes presented in the Results section. The ratio is computed compared to the $\mathcal{O}(n^{2})$ legacy self-attention map size.
We can see the sparsity can reach up to more than $80\%$ while the pairwise computation ratios range from 5 to $35\%$ only.

The complexity of the network is defined by $\mathcal{O}(N(d^{2}(2n-k)+hd)$ where $h\ll n^{2}$ denotes the \emph{fixed} number of computations of the self-attention module.
Acceleration of the proposed Transformer via architectural modification or neural networks acceleration (e.g. pruning, quantization) \cite{lin2021survey} is left for future work.

In comparison, the existing AR-BP SOTA method of \cite{nachmani2021autoregressive} requires $\mathcal{O}(2N(nd_{v}d_{f}d_{g}+n^{2}d_{v} + (n-k)^{2} + d_{f} ))$ more operations than Hyper BP\cite{nachmani2019hyper}, where $d_{v},d_{f},d_{g}$ are the number of variable nodes, complexity of the hyper network and the network respectively.
Typical networks $f$ and $g$ are a 128 dimensional FC network with 4 layers and  a 16 dimensional FC network with 2 layers respectively.{\color{black} With these hyperparameters, the $nd_{v}d_{f}d_{g}$ part of the complexity is approximately $nd_{v}128^{2}$. Moreover, the method of \cite{nachmani2021autoregressive} cannot be parallelized due to the hypernetwork structure, i.e. each sample defines a different set of parameters}.
Thus, our network largely surpasses the existing state-of-the-art error rates with much less time complexity.
{\color{black}
Hyper-BP\cite{nachmani2019hyper} is little less computationally intensive since it adds a complexity term of $\mathcal{O}\left ( 2Nnd_{v} d_f d_g \right )$ on top of BP, but this approach as well as the original model-based network of \cite{nachmani2016learning} are much less competitive in term of accuracy.
}

\section{Conclusions}
We present a novel Transformer architecture for the decoding of algebraic block codes. The proposed model allows effective representation of interactions based on the high dimensional embedding of the channel output and a code-dependent masking scheme of the self-attention module. 
The proposed framework is more efficient than standard Transformers due to the sparse design of existing block codes, making its training and deployment extremely affordable. 
Even with a limited number of computational blocks, it outperforms popular (neural) decoders on a broad range of code families.
We believe this new architecture will allow the development and adoption of new families of performing Transformer based codes capable to set new standards in this field.

\section*{Acknowledgments}
This project has received funding from the European Research Council (ERC) under the European Union's Horizon 2020 research and innovation programme (grant ERC CoG 725974). The contribution of the first author is part of a PhD thesis research conducted at Tel Aviv University.

\bibliographystyle{plain}
\bibliography{references}

\end{document}